\documentclass[sigconf, screen]{acmart} 
\AtBeginDocument{%
  }

\usepackage{multirow}
\usepackage{booktabs}
\usepackage{colortbl}
\usepackage{xcolor}
\usepackage{makecell}
\usepackage{geometry}
\usepackage{arydshln}
\usepackage{graphicx}
\usepackage{subcaption}  
\usepackage{textcomp}
\usepackage{wasysym}

\setcopyright{acmlicensed}
\copyrightyear{2025}
\acmYear{2025}
\acmDOI{XXXXXXX.XXXXXXX}
\acmConference[MM '25]{Proceedings of the 33rd ACM International Conference on Multimedia}{October 27--31,
  2025}{Dublin, Ireland}
\acmISBN{978-1-4503-XXXX-X/2018/06}

\begin{document}

\definecolor{myred}{RGB}{220, 0, 0}
\definecolor{myblue}{RGB}{0, 102, 204}
\renewcommand\theadfont{\bfseries}

\title{RGC-VQA: An Exploration Database for Robotic-Generated Video Quality Assessment}

\author{Jianing Jin}
\email{jinjianing0928@sjtu.edu.cn}
\affiliation{%
  \institution{Shanghai Jiaotong University}
  \city{Shanghai}
  \country{China}
}

\author{Jiangyong Ying}
\email{yingjiangyong@chinatelecom.cn}
\affiliation{%
  \institution{China Telecom}
  \country{China}
  }

\author{Huiyu Duan}
\email{huiyuduan@sjtu.edu.cn}
\affiliation{%
  \institution{Shanghai Jiaotong University}
  \city{Shanghai}
  \country{China}
}

\author{Liu Yang}
\email{ylyl.yl@sjtu.edu.cn}
\affiliation{%
  \institution{Shanghai Jiaotong University}
  \city{Shanghai}
  \country{China}
}

\author{Sijing Wu}
\email{wusijing@sjtu.edu.cn}
\affiliation{%
  \institution{Shanghai Jiaotong University}
  \city{Shanghai}
  \country{China}
}

\author{Yunhao Li}
\email{lyhsjtu@sjtu.edu.cn}
\affiliation{%
  \institution{Shanghai Jiaotong University}
  \city{Shanghai}
  \country{China}
}

\author{Yushuo Zheng}
\email{yushuozheng@sjtu.edu.cn}
\affiliation{%
  \institution{Shanghai Jiaotong University}
  \city{Shanghai}
  \country{China}
}

\author{Xiongkuo Min}
\email{minxiongkuo@sjtu.edu.cn}
\affiliation{%
  \institution{Shanghai Jiaotong University}
  \city{Shanghai}
  \country{China}
}

\author{Guangtao Zhai}
\email{zhaiguangtao@sjtu.edu.cn}
\affiliation{%
  \institution{Shanghai Jiaotong University}
  \city{Shanghai}
  \country{China}
}

\renewcommand{\shortauthors}{Jin et al.}

\begin{abstract}

As camera-equipped robotic platforms become increasingly integrated into daily life, robotic-generated videos have begun to appear on streaming media platforms, enabling us to envision a future where humans and robots coexist. We innovatively propose the concept of \underline{\textbf{R}}obotic-\underline{\textbf{G}}enerated \underline{\textbf{C}}ontent (\textbf{RGC}) to term these videos generated from egocentric perspective of robots. The perceptual quality of RGC videos is critical in human–robot interaction scenarios, and RGC videos exhibit unique distortions and visual requirements that differ markedly from those of professionally-generated content (PGC) videos and user-generated content (UGC) videos. However,
dedicated research on quality assessment of RGC videos is still lacking. To address this gap and to support broader robotic applications, we establish the first \underline{\textbf{R}}obotic-\underline{\textbf{G}}enerated \underline{\textbf{C}}ontent \underline{\textbf{D}}atabase (\textbf{RGCD}), which contains a total of 2,100 videos drawn from three robot categories and sourced from diverse platforms. A subjective VQA experiment is conducted subsequently to assess human visual perception of robotic-generated videos. Finally, we conduct a benchmark experiment to evaluate the performance of 11 state-of-the-art VQA models on our database. Experimental results reveal significant limitations in existing VQA models when applied to complex, robotic-generated content, highlighting a critical need for RGC-specific VQA models. Our RGCD is publicly available at: \url{https://github.com/IntMeGroup/RGC-VQA}.
   
\end{abstract}



\begin{teaserfigure}
    \centering
    \vspace{2mm}
    \includegraphics[width=\linewidth]{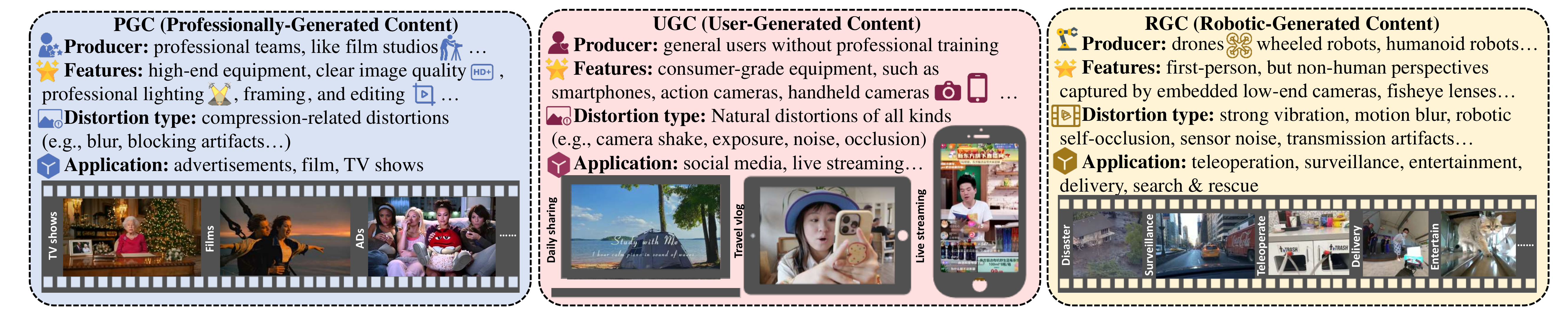} 
    \vspace{1mm}
    \caption{We initiate the concept of Robot-Generated Content (RGC), referring to egocentric video captured by robots operating autonomously or under remote control. Unlike Professionally-Generated Content (PGC) and User-Generated Content (UGC), RGC exhibits fundamental differences in terms of content producers, inherent features, distortion types, and application scenarios. These distinctions present unique challenges in the study of video quality assessment.}
    \label{fig:teaser}
    \vspace{5mm}
\end{teaserfigure}


\maketitle

\section{Introduction}

Recent years have witnessed the rapid development of robotics and intelligent machines. Autonomous and semi-autonomous devices, such as vehicles, drones, and household cleaning robots, have become increasingly embedded into everyday life. Most of these machines are equipped with cameras that continuously generate vast quantities of visual data \cite{CCD,bdd100k,Ford-Campus,JAAD,awesome-slam-datasets}. Concurrently, research on robotic manipulators and humanoid robots has generated extensive egocentric visual data captured from the egocentric view of robots\cite{ario,droid,fourier,unitree,open-x-embodiment,rh20t}.
Given the uniqueness of sources and visual characteristics of these robot captured videos, we propose to term them \underline{\textit{R}}obotic-\underline{\textit{G}}enerated \underline{\textit{C}}ontent (\textbf{RGC}), thereby distinguishing them from the two conventional categories of video content: professionally-generated content (PGC), which are produced by experts using high-end equipment, and user-generated content (UGC), which are captured by amateurs with consumer-grade handheld devices.  

Numerous studies have developed video quality assessment (VQA) databases for PGC and UGC videos, focusing on either synthetic distortions \cite{seshadrinathan2010study,de2010h,moorthy2012video,nuutinen2016cvd2014,wang2016mcl} or authentic distortions \cite{ghadiyaram2017capture,hosu2017konstanz,sinno2018large,wang2019youtube,ying2021patch}, and more recent datasets aim to combine both \cite{li2020ugc,yu2021predicting,wang2021rich,zhang2023md,icme21}. Previous studies have investigated the quality assessment of embodied images \cite{EPD,li1} and shifted the focus from solely human perception to include machine preference \cite{li2}. However, no research has specifically addressed the quality assessment of RGC videos. RGC videos play a critical role in a wide range of real-world applications, such as enabling human-in-the-loop robot control, facilitating inspection of hard-to-reach environments, and supporting field research, \textit{etc.}, where Quality of Experience (QoE) of human viewers is of great importance. However, as RGC videos are sourced from cameras fixed on moving robots, offering machine-centric viewpoints to facilitate human oversight, they exhibit distinct visual content and unique quality issues that differ markedly from those of PGC or UGC videos. Moreover, the uniqueness of RGC applications impose distinct visual quality requirements. Therefore, dedicated research on RGC video quality assessment is of urgent need.

To facilitate the study of perceptual quality in egocentric robot-view videos and support broader robotic applications, we introduce the concept of \underline{\textit{R}}obotic-\underline{\textit{G}}enerated \underline{\textit{C}}ontent (RGC) and present the first database for \underline{\textit{R}}obotic-\underline{\textit{G}}enerated \underline{\textit{C}}ontent video \underline{\textit{D}}atabase, termed \textbf{RGCD}. Specifically, 2,100 RGC videos generated by three kinds of robotic devices: wheeled robots, drones, and robotic arms, are collected from various sources, including both online video platforms and academic datasets. Based on the collected videos, we further conduct an in lab subjective experiment to study perceptual quality of robotic-generated videos. Overall, RGCD includes \textit{2,100} mean opinion scores (MOSs) produced from over \textit{31K} quality ratings.
Based on the RGCD database, we conduct a comprehensive benchmark experiment to evaluate the performance of 11 state-of-the-art VQA models on RGC video quality assessment. Experimental results reveal a substantial gap between human perception and current VQA model predictions on RGC videos, indicating that existing VQA methods are not well suited for predicting human visual perception of RGC videos. The main highlights of this work are fourfold:

\begin{itemize}
\item We introduce the concept of Robot-Generated Content (RGC), referring to video content captured from the egocentric perspective of robots operating autonomously or under remote control.
\item We establish RGCD, the first robotic-generated content video quality assessment database that contains 2100 RGC videos with corresponding over 31K subjective quality ratings.
\item We conduct a systematic analysis of perceptual quality of RGC videos based on the constructed database.
\item A extensive benchmark experiment is conducted to evaluate the performance of 11 state-of-the-art VQA models on perceptual quality assessment of RGC videos.
\end{itemize}

\begin{figure*}[t]
    \centering
    \includegraphics[width=1\linewidth]{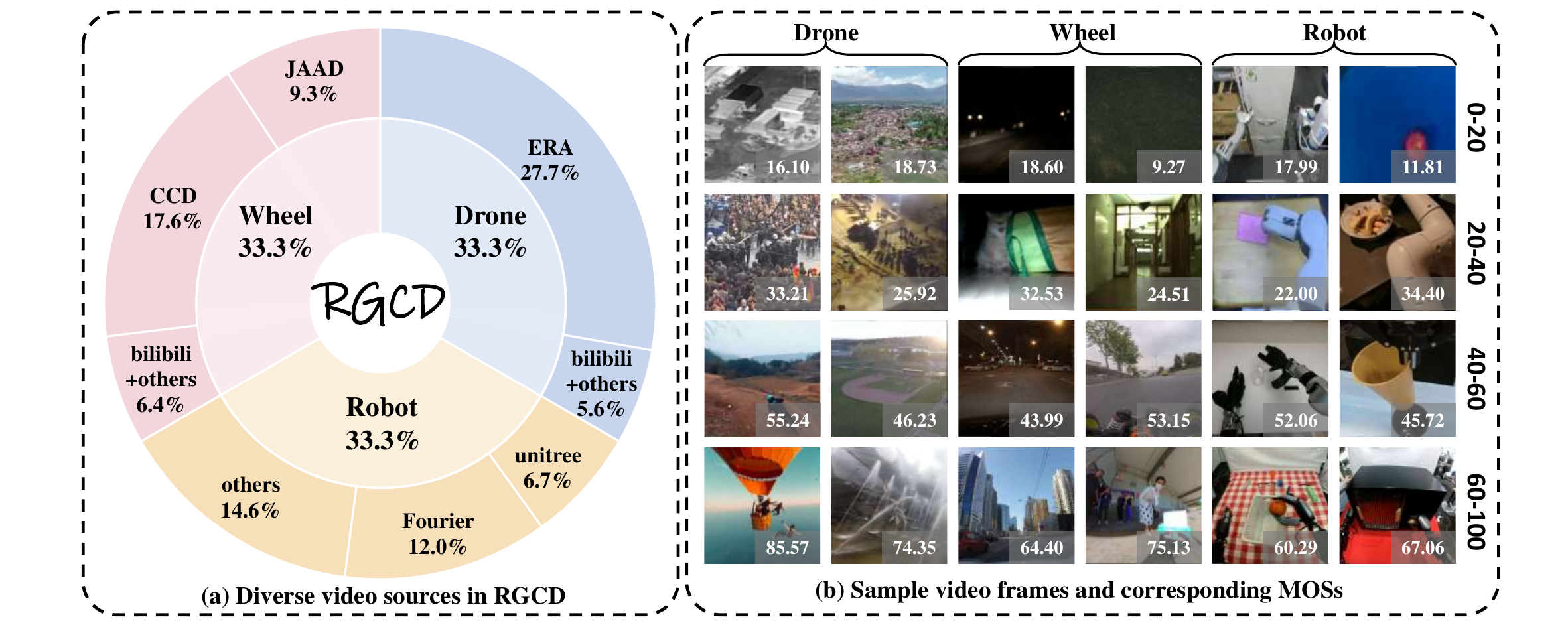}
    \caption{We establish the first Robot-Generated Content Database (RGCD) to study the RGC quality assessment problem. The database consists of 2100 videos from diverse sources and are categorized into three classes:  drones, wheeled robots, and robotic arms, with 700 videos in each category (a). High-quality MOS annotations are provided for each video, and at the same time, each category encompasses a wide range of quality levels (b).}
    \label{fig:RGCD}
    \vspace{2mm}
\end{figure*}

\section{Related Work}

\subsection{VQA Databases}

Video quality assessment (VQA) databases have evolved significantly to address diverse content types and distortion scenarios. Early databases primarily focused on synthetic distortions (\textit{e.g.}, compression artifacts, noise) due to limited source videos \cite{seshadrinathan2010study,de2010h,moorthy2012video,nuutinen2016cvd2014,wang2016mcl}. With the rise of user-generated content (UGC), datasets like \cite{nuutinen2016cvd2014,ghadiyaram2017capture,hosu2017konstanz,sinno2018large,wang2019youtube,ying2021patch} introduced authentic distortions (\textit{e.g.}, motion blur, exposure changes) from real-world recordings. Recent efforts \cite{li2020ugc,yu2021predicting,wang2021rich,zhang2023md,icme21} combine both synthetic and authentic distortions, while specialized datasets target emerging formats like short-form videos \cite{lu2024kvq}. However, none of these databases address Robot-Generated Content (RGC), which exhibits unique distortions from machine-captured perspectives (\textit{e.g.}, mechanical vibrations, egocentric motion patterns). Our RGCD database fills this gap by providing 2,100 RGC videos with MOS annotations, covering diverse machine types (drones, wheels, robots) and in-the-wild scenarios.

\subsection{VQA methods}

Traditional VQA methods rely on handcrafted features (\textit{e.g.}, NIQE \cite{niqe}, TLVQM \cite{korhonen2019two}) but struggle with complex real-world distortions. Deep learning-based approaches (\textit{e.g.}, VSFA~\cite{li2019quality}, SimpleVQA~\cite{sun2022deep}) leverage pretrained networks to extract semantic features, achieving state-of-the-art performance on UGC datasets. Recent advances like FAST-VQA~\cite{wu2022fast} and DOVER~\cite{wu2023exploring} further optimize for efficiency and temporal modeling. However, their generalization performance on robot-generated content, as in our RGCD, remains uncertain. Currently, there are no models specifically tailored for RGC, highlighting a gap in existing VQA research and the need for domain-adaptive solutions.

\subsection{Robotics and Embodied Intelligence}

The recent surge in embodied AI has fostered large-scale efforts to unify robot perception and control across diverse embodiments \cite{open-x-embodiment, ario}. X-Embodiment \cite{open-x-embodiment} presents a cross-robot learning framework using 60 datasets from 22 embodiments. Meanwhile, the rapid development of humanoid and quadruped robots, such as Unitree \cite{unitree}, Tesla Optimus, and Boston Dynamics' Atlas, is pushing robots from lab environments into real-world deployments. These robots often operate in dynamic, unstructured environments while relying heavily on egocentric visual feedback for navigation, manipulation, and interaction \cite{droid, fourier, unitree}. Prior work has explored the quality assessment of embodied images \cite{EPD, li1}, expanding the scope of evaluation from human to machine preferences \cite{li2}. However, none of the existing datasets or benchmarks systematically evaluate the perceptual quality of robotic-generated content (RGC), especially in the video domain.  Our work lays the foundation for the content produced by robots and  embodied intelligence.

\section{Database Construction and Analysis}
\begin{figure*}[t]
  \centering
  \vspace{1mm}
  \begin{subfigure}{0.24\textwidth}
    \includegraphics[width=\linewidth]{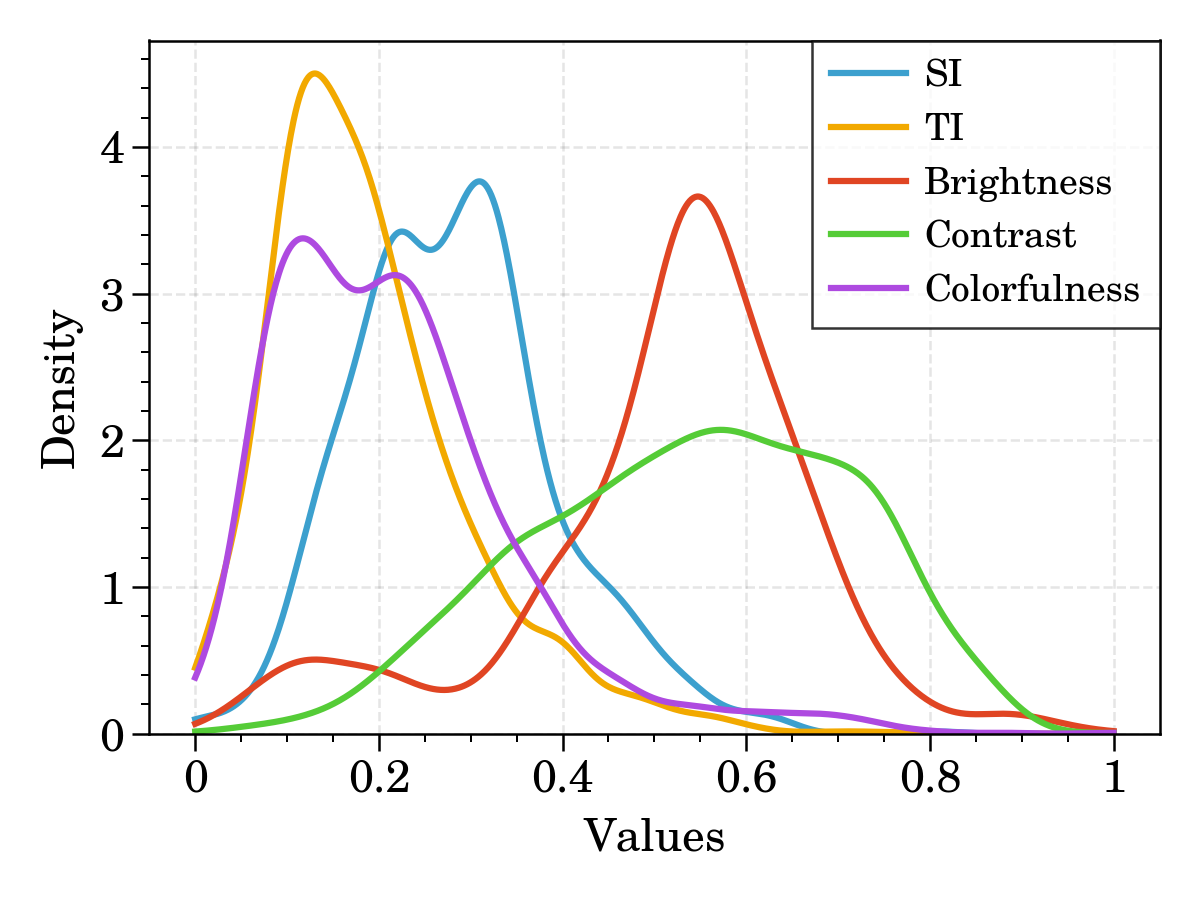}
    \caption{All}
    \label{siti:all}
  \end{subfigure}
  \hfill
  \begin{subfigure}{0.24\textwidth}
    \includegraphics[width=\linewidth]{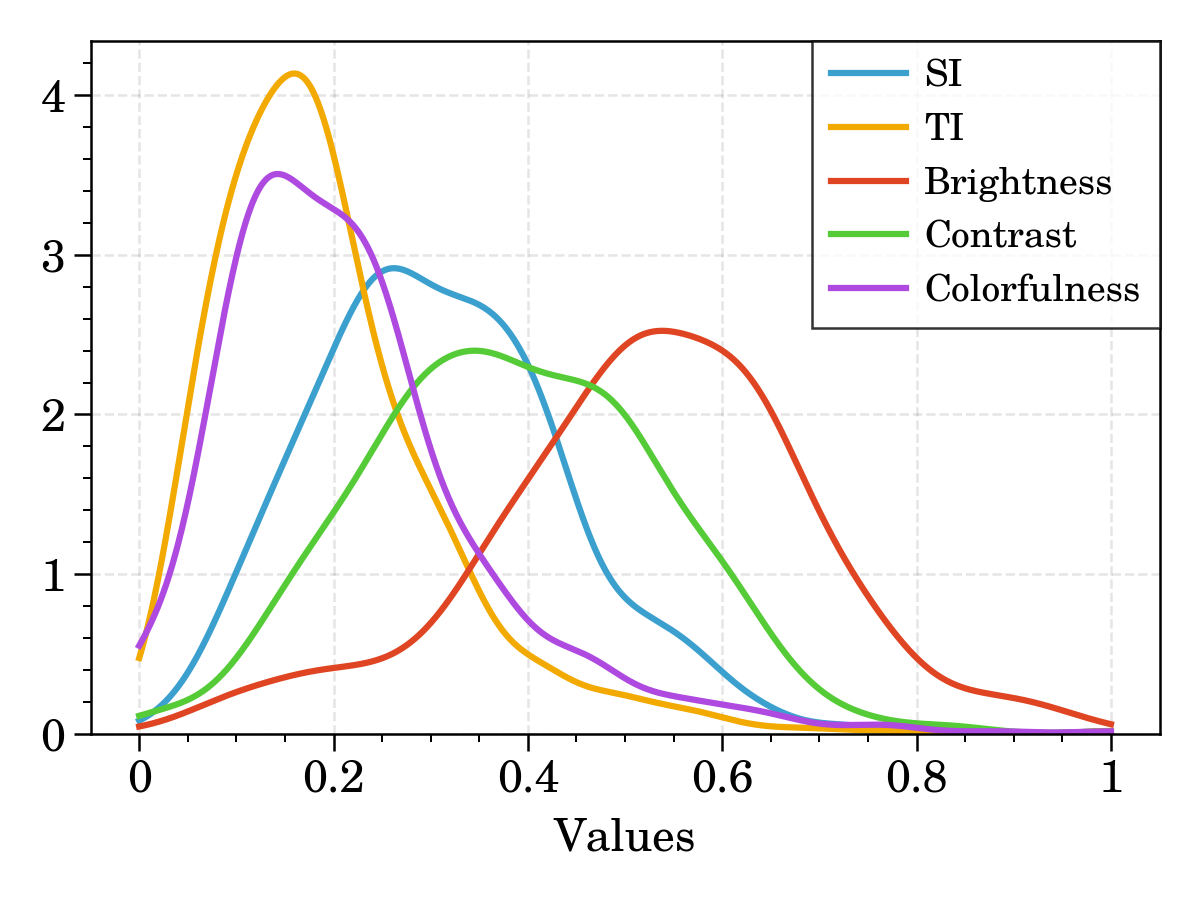}
    \caption{Drone}
    \label{siti:drone}
  \end{subfigure}
  \hfill
  \begin{subfigure}{0.24\textwidth}
    \includegraphics[width=\linewidth]{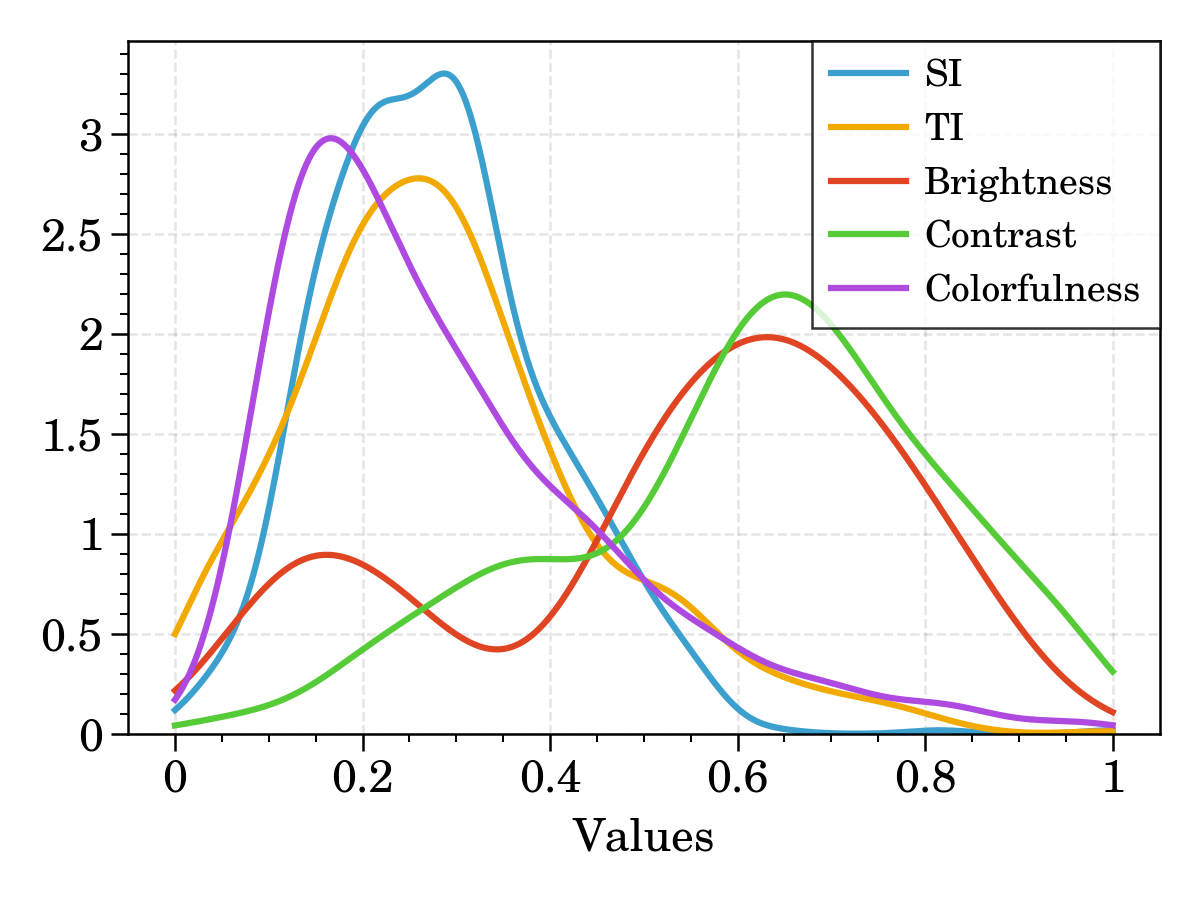}
    \caption{Wheel}
    \label{siti:wheel}
  \end{subfigure}
  \hfill
  \begin{subfigure}{0.24\textwidth}
    \includegraphics[width=\linewidth]{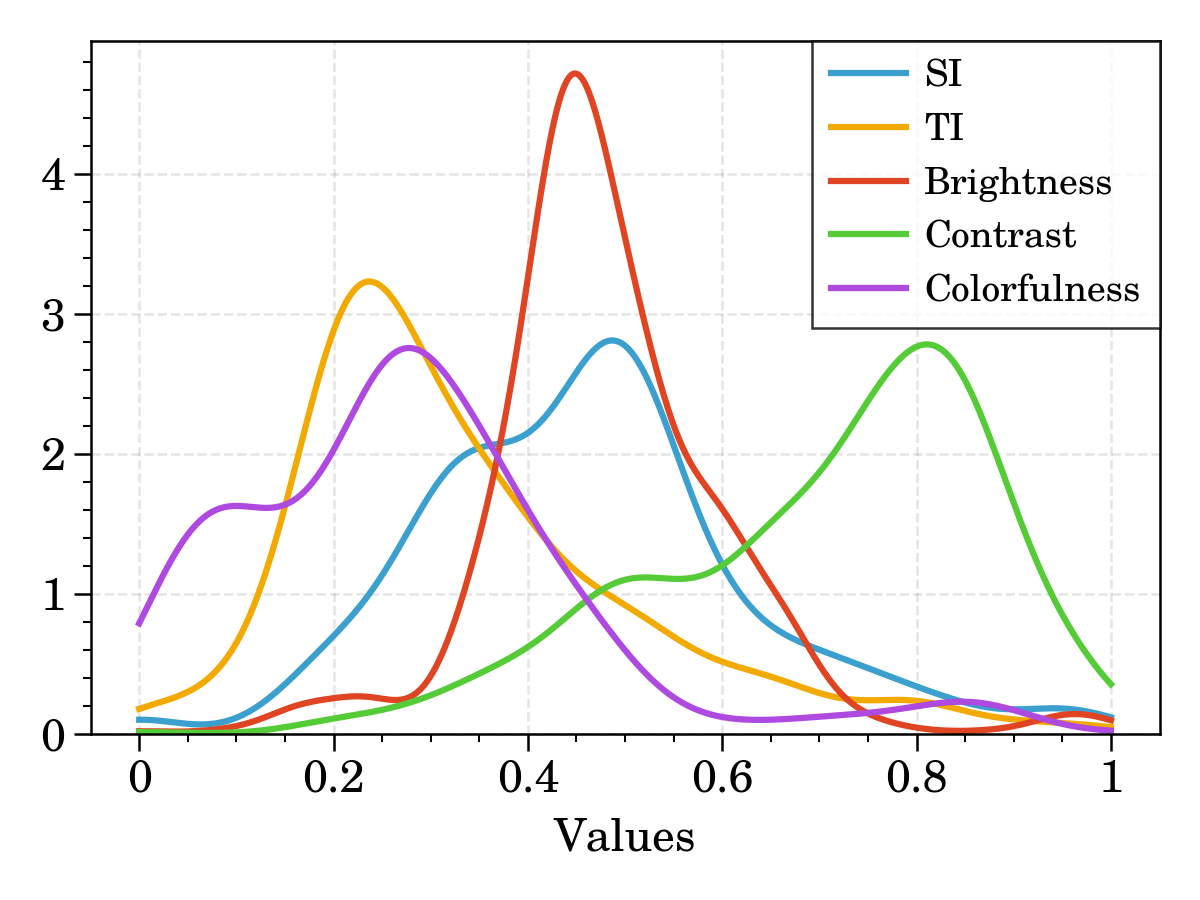}
    \caption{Robot}
    \label{siti:robot}
  \end{subfigure}
  \caption{The feature distributions of RGCD in terms of different categories: (a) the full database, (b) drone, (c) wheeled robot, (d) humanoid robot.}
  \label{fig:siti}
  \vspace{1mm}
\end{figure*}

\begin{figure*}[t]
  \centering
  \vspace{1mm}
  \begin{subfigure}{0.24\textwidth}
    \includegraphics[width=\linewidth]{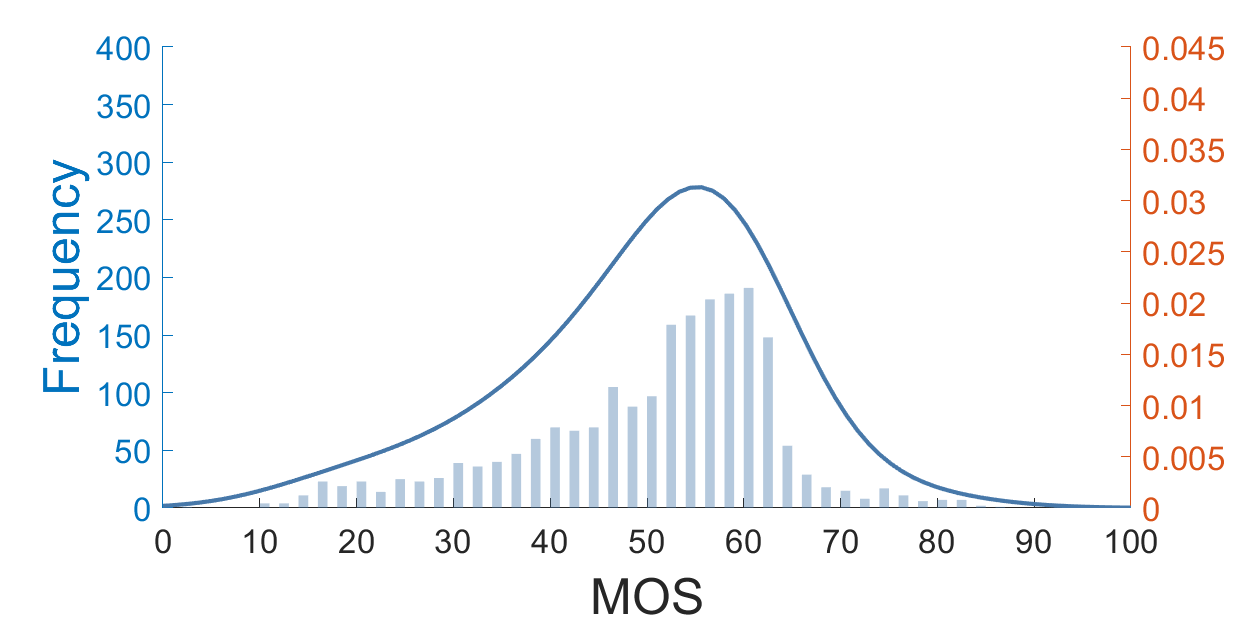}
    \caption{All}
    \label{mos:all}
  \end{subfigure}
  \hfill
  \begin{subfigure}{0.24\textwidth}
    \includegraphics[width=\linewidth]{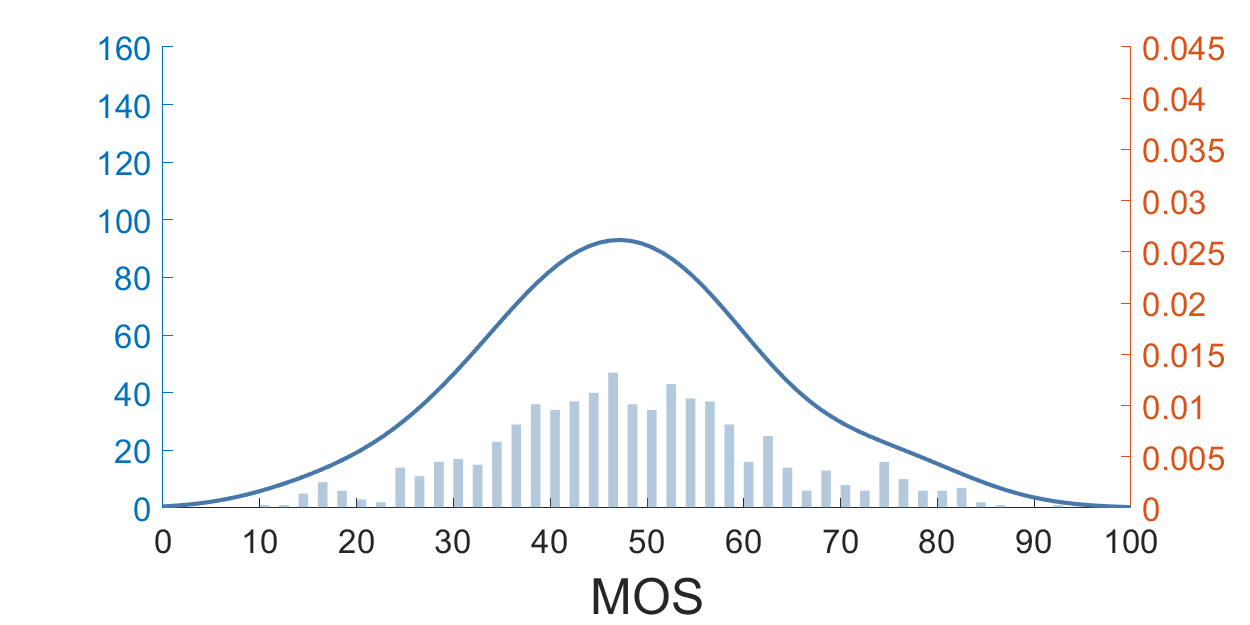}
    \caption{Drone}
    \label{mos:drone}
  \end{subfigure}
  \hfill
  \begin{subfigure}{0.24\textwidth}
    \includegraphics[width=\linewidth]{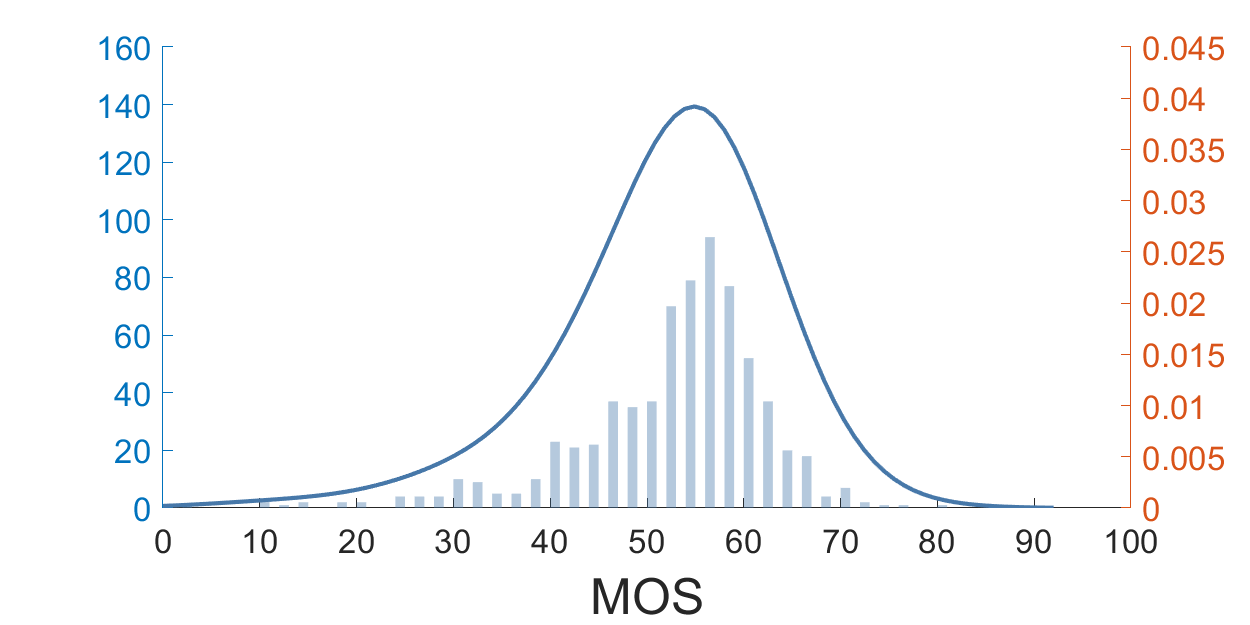}
    \caption{Wheel}
    \label{mos:wheel}
  \end{subfigure}
  \hfill
  \begin{subfigure}{0.24\textwidth}
    \includegraphics[width=\linewidth]{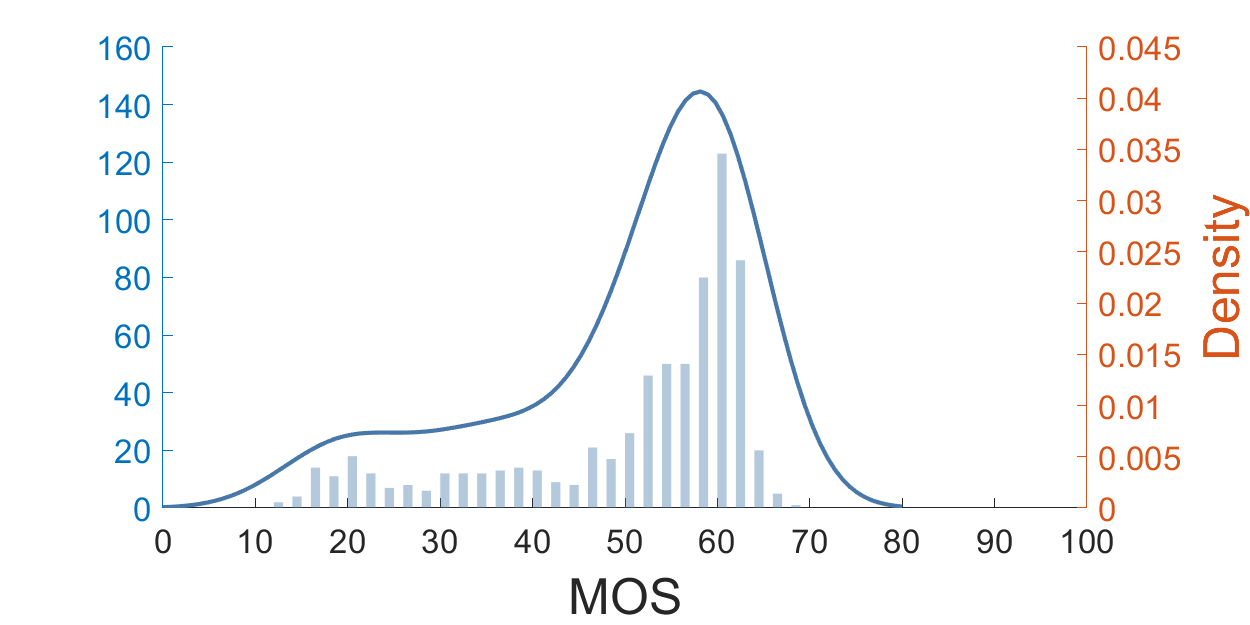}
    \caption{Robot}
    \label{mos:robot}
  \end{subfigure}
  \caption{The MOS distributions of RGCD in terms of different categories: (a) the full database, (b) drone, (c) wheeled robot, (d) humanoid robot.}
  \label{fig:mos}
  \vspace{2mm}
\end{figure*}

In this section, we introduce \textbf{RGCD}, the first diverse robotic-generated content video quality assessment database, designed to address the lack of a comprehensive video dataset captured from machine perspectives and to study perceptual quality of these videos. The database includes 2,100 videos featuring various scenarios and is accompanied by over 31K high-quality human opinion ratings, as shown in Figure~\ref{fig:RGCD}. Below, we detail the construction procedure of the RGCD database and present an in-depth analysis of its content.

\subsection{RGC Video Collection}
\label{sec:3.1}
The main principle of the RGC video collection process is to cover a wide variety of robot types and in-the-wild scenarios, while ensuring that the selected viewpoints authentically reflect the egocentric perspective of machines.
Through a comprehensive search of existing academic datasets, we found that several research domains, such as embodied intelligence, SLAM, and autonomous driving, contain abundant RGB video data captured from the first-person viewpoint of robots. As a result, we broadly categorize the collected videos into three groups: drones, wheeled robots, and humanoid robots.

For drones and wheeled robots, we began with the Awesome-SLAM-Dataset \cite{awesome-slam-datasets}, from which we selected sequences that contain RGB video data, including agricultural robots \cite{agri-robot}, as well as indoor and outdoor mapping robots. We further extended our collection to cover a variety of drone and autonomous driving datasets designed for diverse tasks. Specifically, drone data are obtained from \cite{Zurich, VIRAT, UZH-FPV, EuRoC, ERA}, while driving datasets are sourced from \cite{bdd100k, Ford-Campus, CCD, KITTI, JAAD}.

For humanoid robots, we extracted wrist-camera and other egocentric views from datasets such as X-Embodiment \cite{open-x-embodiment} and ARIO \cite{ario}. However, data collected from identical robotic platform often involve repetitive executions of the same tasks, leading to nearly indistinguishable motion patterns and views. To enhance the diversity of our dataset, we selected samples from a wide range of scenarios, limiting each unique scene to a maximum of five repetitions. Meanwhile, we also incorporated more in-the-wild datasets such as \cite{fourier} and \cite{unitree}, which more closely reflect the real-world egocentric perspectives of Robot-Generated Content (RGC) as perceived by humans.

Unfortunately, most of the existing datasets in this category are restricted to robotic arm operations. We were unable to locate large-scale egocentric video data captured during locomotion of humanoid or quadruped robots. Instead, many current researches simulate such perspectives using human actors, as seen in datasets like Ego4D \cite{ego4d}. We sincerely look forward to more open-source datasets providing authentic first-person visual data from mobile robots.

In addition to academic sources, we also collected nearly 200 videos from the popular streaming platform, Bilibili \cite{bilibili}, mainly featuring drones and wheeled robots. These genuine RGC videos include content captured from remote-controlled toy-cars, drones (including FPV drones), and sweeping robots. These early forms of entertainment-oriented robot-generated content include but are not limited to aerial views in various outdoor scenes, remote-controlled vehicles picking up deliveries, and sweeping robots interacting with pets. To ensure consistency, all collected videos from streaming platforms are processed by cropping to remove watermarks. 

All videos in RGCD are trimmed to a duration of 4 to 12 seconds, aiming to retain representativeness while eliminating redundancy.In total, we compiled a dataset of 2,100 RGC video clips, evenly divided across the three categories, with 700 videos in each.

\subsection{Subjective Experiment}
\label{sec:3.2}
We propose to assess the perceptual quality of RGC videos using a single overall score. Specifically, subjects evaluate each video based on color fidelity, noise levels, visible artifacts, and temporal consistency, then provide a overall quality rating.

A in-lab subjective experiment is conducted to measure the perceptual quality of robotic-generated videos. To ensure the quality of the experiment results, different from the crowd-sourced method adopted by previous VQA database, we recruit 15 college students as subjects and conduct the experiment in a lab environment following the recommendation of ITU-BT.500 \cite{series2012methodology}. The videos are displayed on a screen with a resolution of 3840$\times$2160, and the viewers are seated at a distance of about 2 feet from the screen and about 5 feet from each other. 

The subjective experiment is conducted on the graphical user interface (GUI), where one screen is used to display the videos, and another display the rating panel for participants to record their scores and switch between videos.  Following the single stimulus absolute category rating (SSACR) method, subjects are instructed to rate the perceptual quality for each video on a continuous scale ranging from 0 to 5, with a resolution of one decimal place (\textit{i.e.}, 0.0 to 5.0), allowing for fine-grained quality assessment. Before the subjective experiments, we train the subjects with detailed rating criteria and numerous examples from each quality level and each type of distortions. Then, a test session with about 20 videos is conducted to familiarize the subjects with the overall quality range of the dataset. To ensure unbiased evaluation, the order of the video playlist is randomized for every subject.  Finally, we collect a total of 31500 quality ratings (15 participants $\times$ 2100 ratings), for 2100 scores each participant.

\subsection{Subjective Data Processing}
\label{sec:3.3}

\begin{table*}[t]
\caption{Performance of 11 state-of-the-art models and zero-shot performance of 5 DNN-based models on our overall RGCD database and three categories. $\clubsuit$, $\diamondsuit$, and $\spadesuit$ denote traditional IQA methods, traditional VQA models and deep-learning-based VQA models, respectively. * indicates models trained and fine-tuned with separate weights for each categories and the entire RGCD database. The best results and the second-best results are highlighted in \textcolor{myred}{\textbf {red}} and \textcolor{myblue}{\textbf {blue}}, respectively.}
\label{big table}
\centering
\begin{tabular}{l|ccc|ccc|ccc|ccc}
\toprule
\textbf{Categories} & 
\multicolumn{3}{c|}{\textbf{Drone}} & 
\multicolumn{3}{c|}{\textbf{Wheel}} & 
\multicolumn{3}{c|}{\textbf{Robot}}  &
\multicolumn{3}{c}{\textbf{All}} \\
\textbf{Methods / Metrics} & SRCC & KRCC & PLCC & SRCC & KRCC & PLCC & SRCC & KRCC & PLCC 
& SRCC & KRCC & PLCC  \\
\midrule
$\clubsuit$NIQE\cite{niqe}  & 0.4335 & 0.3094 & 0.3813 & 0.5751 & 0.4072 & 0.6445 & 0.6546 & 0.4716 & 0.7353 & 0.4987 & 0.3485 & 0.4835\\
$\clubsuit$QAC\cite{qac}  & -0.0889 & -0.0648 & 0.0282 & 0.0230 & 0.0152 & -0.0478 & -0.1571 & -0.1058 & 0.0598 & 0.0105 & 0.0065 & 0.0568 \\
$\clubsuit$HOSA\cite{hosa}  & 0.3746& 0.2731 & 0.3299 & 0.5890 & 0.4219 & 0.6180 & 0.4994 & 0.3466 & 0.3645 & 0.4286 & 0.3033 & 0.3911 \\
$\diamondsuit$TLVQM\cite{korhonen2019two}  & 0.8181 & 0.6343 & \textcolor{myblue}{\textbf {0.8745}} & 0.6875 & 0.5187 & 0.7609 & 0.8728 & 0.7118 & \textcolor{myred}{\textbf {0.9708}} & 0.8636 & \textcolor{myblue}{\textbf {0.6840}} & \textcolor{myblue}{\textbf {0.8889}} \\
$\diamondsuit$VIDEVAL\cite{tu2021ugc}  & 0.7412 & 0.5655 & 0.8158 & 0.6987 & 0.5178 & 0.7454 & 0.8795 & 0.7189  & 0.9151 & 0.8141 & 0.6311 & 0.8327\\
$\diamondsuit$RAPIQUE~\cite{tu2021rapique}  & 0.7238 & 0.5491 & 0.8366 & 0.7479 & 0.5630 & 0.8100 & 0.8865 & 0.7157 & 0.8997 & 0.8128 & 0.6216 & 0.7920\\
\hdashline
$\spadesuit$VSFA~\cite{li2019quality} & 0.5127 & 0.3714 & 0.4895 & 0.6649 & 0.4961 & 0.6617 & 0.2479 & 0.1744 & 0.0751 & 0.3663 & 0.2540 & 0.2876 \\
$\spadesuit$GSTVQA~\cite{chen2021learning} & 0.4679 & 0.3256 & 0.4327 & 0.7429 & 0.5357 & 0.6732 & 0.7538 & 0.5404 & 0.7375 & 0.6390 & 0.4526 & 0.5994 \\
$\spadesuit$SimpleVQA~\cite{sun2022deep}  & 0.7140 & 0.5266 & 0.6984 & 0.7843 & 0.5873 & 0.7697 & 0.8856 & 0.7046 & 0.9083 & 0.7787 & 0.5848 & 0.7772 \\
$\spadesuit$FAST-VQA~\cite{wu2022fast} & 0.7247 & 0.5365 & 0.7169 & 0.8049 & 0.6117 & 0.7834 & 0.8453 & 0.6539 & 0.8674 & 0.7646 & 0.5708 & 0.7788 \\
$\spadesuit$DOVER~\cite{wu2023exploring} & 0.7453 & 0.5572 & 0.7398 & 0.7969 & 0.6140 & 0.7283 & 0.8214 & 0.6111 & 0.8551 & 0.7725 & 0.5705 & 0.7391 \\
\hdashline
$\spadesuit$VSFA~\cite{li2019quality}*  & 0.8129 & 0.6356 & 0.8323 & 0.8054 & 0.6208 & \textcolor{myblue}{\textbf {0.8446}} & \textcolor{myblue}{\textbf {0.8969}} & \textcolor{myblue}{\textbf {0.7285}} & 0.9480 & 0.8609 & 0.6836 & 0.8815 \\
$\spadesuit$GSTVQA~\cite{chen2021learning}*  & 0.7359 & 0.5392 & 0.6677 & 0.7792 & 0.5846 & 0.7863 & 0.8733 & 0.6978 & 0.9277 & 0.8608 & 0.6780 & 0.8612 \\
$\spadesuit$SimpleVQA~\cite{sun2022deep}*  & \textcolor{myblue}{\textbf {0.8493}} & \textcolor{myblue}{\textbf {0.6635}} & 0.8696 & \textcolor{myred}{\textbf {0.8297}} & \textcolor{myred}{\textbf {0.6479}} & \textcolor{myred}{\textbf {0.8533}} & 0.8911 & 0.7239 & \textcolor{myblue}{\textbf {0.9587}} & \textcolor{myred}{\textbf {0.8785}} & \textcolor{myred}{\textbf {0.7022}} & \textcolor{myred}{\textbf {0.8990}} \\
$\spadesuit$FAST-VQA~\cite{wu2022fast}*  & \textcolor{myred}{\textbf {0.8579}} & \textcolor{myred}{\textbf {0.6843}} & \textcolor{myred}{\textbf {0.8821}} & 0.8218 & 0.6347 & 0.7616 & \textcolor{myred}{\textbf {0.9040}} & \textcolor{myred}{\textbf {0.7360}} & 0.9241 & \textcolor{myblue}{\textbf {0.8658}} & 0.6754 & 0.8468 \\
$\spadesuit$DOVER~\cite{wu2023exploring}*  & 0.7679 & 0.5789 & 0.7681 & \textcolor{myblue}{\textbf {0.8281}} & \textcolor{myblue}{\textbf {0.6414}} & 0.7869 & 0.8480 & 0.6465 & 0.8370 & 0.7891 & 0.5878 & 0.7857 \\
\bottomrule
\end{tabular}
\end{table*}

First, we follow the suggestion of the subjective data processing guidelines recommended by ITU \cite{series2012methodology} to perform outlier detection and subject rejection. The result shows that none of the 15 participants is regarded as an outlier and rejected. Each video has at least 13 valid ratings, and about 2.46\% of the total subjective scores are identified as outliers and are subsequently excluded. For the remaining valid subjective evaluations, we convert these raw ratings into Z-scores by:
\begin{equation}
  z_{ij} = \frac{r_{ij}-\mu_{i}}{\sigma_i}
\end{equation}
where $r_{ij}$ is the original score given by the $i$-th subject to the $j$-th RGC video, $\mu_i$ and $\sigma_i$ respectively represent the mean rating and the standard deviation given by the $i$-th subject. Under the assumption that the Z-scores of a subject follow a standard Gaussian distribution within the range of $\left[-3, +3\right]$, z-scores are linearly scaled to the range of $[0,100]$ as follows:
\begin{equation}
  z’_{ij} = \frac{100(z_{ij}+3)}{6}
\end{equation}
Lastly, the rescaled z-scores $r'_{ij}$ are averaged over subjects to obtain the final mean opinion scores (MOSs), which is formulated as:
\begin{equation}
  \text{MOS}_j=\frac{1}{N}\sum_{i=1}^{N} z'_{ij}{\sigma_i}
\end{equation}
where $N$ is the total number of subjects and $j$ represents the $j$-th RGC video.

\subsection{Data Analysis}
\label{sec:3.4}
\subsubsection{Statistic analysis for RGCD database}
We conduct statistical analysis for our full RGCD
database and three categories in terms of five low-level vision feature dimensions including: colorfulness, brightness, contrast, spatial information (SI) and temporal information (TI). As shown in Figure \ref{fig:siti}, the constructed RGCD database exhibits broad feature characteristics across five video quality related features both for the full database and the three categories. It can be observed that the majority of features
span a wide range of normalized values, indicating the feature diversity inherent in out database.
Additionally, the SI and TI also exhibit broad value distributions, suggesting that the videos in RGCD posse both spatial and temporal richness.

\subsubsection{Perceptual quality analysis for RGC videos}
Based on the collected MOS data, we conduct a systematic analysis of perceptual quality across the full database and three distinct categories (drone, wheeled robot, and humanoid robot). Figure \ref{mos:all} presents the MOS distribution for the entire dataset, revealing a broad coverage of scores, which indicates diverse perceptual quality levels among the samples. To investigate category-specific characteristics, we further visualize the MOS distributions for each individual category in Figure \ref{mos:drone}-\ref{mos:robot}. Notably, while the four distributions share similarities, distinct patterns emerge in detail. For instance, The drone category demonstrates a wider dispersion, implying greater variability in the database. Conversely, the wheel category exhibits a tighter score clustering around the middle MOS values (55–65 range), suggesting more consistent quality perception, but also indicating a lack of higher-quality content in this category, as evidenced by the absence of scores in the upper MOS range. This phenomenon become more serious when it comes to the robot category. The overall perceptual quality is relatively lower, which is the main reason of the abrupt drop in frequency around the score of 65 in the MOS distribution of the full database. In addition, the comprehensive coverage of diverse robot types and in-the-wild scenarios underscores the richness and representativeness of our RGC video database.

\section{Experiments}

\subsection{Experimental Setup}

\subsubsection{Baseline Methods} 
Since there is no method specialized for robot-generated content quality assessment task, we benchmark 11 state-of-the-art models on our RGCD dataset. These methods are grouped into two categories: (1) traditional image quality assessment (IQA) and video quality assessment (VQA) methods, including NIQE \cite{niqe}, QAC \cite{qac}, HOSA \cite{hosa}, TLVQM \cite{korhonen2019two}, VIDEVAL \cite{tu2021ugc}, and RAPIQUE ~\cite{tu2021rapique}; and (2) deep learning-based VQA models, encompassing VSFA~\cite{li2019quality}, GSTVQA~\cite{chen2021learning}, SimpleVQA~\cite{sun2022deep}, FAST-VQA~\cite{wu2022fast}, DOVER~\cite{wu2023exploring}. To ensure fair comparison, all DNN-based models are evaluated using their official implementations and default configurations.
Evaluating these models on the RGCD dataset is particularly important, as RGC videos contain complex motion patterns, unstable viewpoints, and diverse content, which pose new challenges beyond those in existing VQA datasets. This makes RGCD a valuable benchmark for testing the robustness and generalization ability of current VQA techniques.

\subsubsection{Evaluation Metrics} 
To evaluate the relationship between predicted scores and MOS, we adopt four commonly used metrics: Spearman Rank Correlation Coefficient (SRCC), Pearson Linear Correlation Coefficient (PLCC), and Kendall’s Rank Correlation Coefficient (KRCC). Among these, SRCC measures the strength and direction of the monotonic relationship, while KRCC provides a more conservative measure of monotonicity and is more robust. PLCC evaluates the linear relationship between predicted scores and MOS.

\subsubsection{Implementation Details} 
We follow a standard 4:1 ratio to split the data into training and test sets for each individual category, as well as for the entire dataset. This strategy ensures that all three categories are proportionally represented. Such balanced distribution helps to mitigate potential evaluation bias and supports fair performance comparison across different model types. Other training parameters are consistent with those in the officially released versions.

\subsection{Evaluation of VQA Models on RGCD}

We first evaluate the performance of 11 state-of-the-art video quality assessment (VQA) models on our established RGCD database, as summarized in Table~\ref{big table}. The results show that traditional image quality assessment (IQA) methods generally perform poorly across all categories, particularly due to their inability to capture temporal dynamics, which are critical in video quality assessment. Among them, NIQE \cite{niqe} achieves moderate performance on the robot category but fails to provide reliable predictions on drone and wheel videos.Compared to IQA methods, traditional VQA models significantly better. Notably, TLVQM \cite{korhonen2019two} reaches a PLCC of 0.9708 on the robot category, indicating strong alignment with human judgments in that domain. 

DNN-based methods demonstrate generally stronger performance, validating the effectiveness of deep temporal and spatial feature modeling. SimpleVQA~\cite{sun2022deep} and FAST-VQA~\cite{wu2022fast} dominate most metrics, suggesting their robustness and strong generalization capability. Notably, DOVER~\cite{wu2023exploring} exhibits relatively weaker performance compared to other deep-learning-based models. This is likely because DOVER introduces a aesthetic factor, whereas robot-egocentric videos prioritize task completion over artistic framing. However, we observe a performance gap between categories. Surprisingly, the robot category tends to achieve higher correlations, possibly due to the more noticeable motion jitters in its content, which make quality differences more distinguishable. In contrast, videos in the drone and wheel categories are more challenging, likely due to fast camera movements and diverse visual contexts. These results reveal the limitations of existing VQA models in handling complex robotic video content, indicating a gap in their robustness and adaptability. Our proposed RGCD dataset addresses this challenge by offering a broad spectrum of real-world robotic scenarios, serving as a valuable resource to expose current model weaknesses and drive the development of RGC-tailored VQA approaches.

\subsection{Zero-shot Performance of Pretrained DNN-Based VQA Models}

To further examine the generalization ability of existing DNN-based VQA models, as shown in Table~\ref{big table}, we evaluate five state-of-the-art pretrained models on our RGCD dataset without fine-tuning. The results show a clear performance gap across models, reflecting varying robustness when transferred to robotic content. Some models, like SimpleVQA ~\cite{sun2022deep} and FAST-VQA ~\cite{wu2022fast}, demonstrate better generalization across categories, while others, such as VSFA~\cite{li2019quality}, struggle, especially on challenging robot videos. DOVER~\cite{wu2023exploring} shows promising results on drone content, likely due to its temporal-aware design, and GSTVQA~\cite{chen2021learning} exhibits moderate stability across different categories. These findings highlight the limitations of models trained on generic datasets when applied to robotic scenarios. Our RGCD dataset thus provides a valuable benchmark for identifying generalization weaknesses and advancing more robust VQA methods tailored to real-world robotics.

\section{Conclusion}

In this work, we pioneer the study of Robot-Generated Content (RGC), addressing a critical gap as machines like drones, robotic arms, and humanoid robots become pervasive in daily life and media production. Specifically, we introduce the first large-scale RGC database, RGCD, which consists of 2100 videos with mean opinion scores (MOS) annotations, collected from diverse sources to reflect real-world RGC scenarios. Our comprehensive benchmarking of 11 state-of-the-art VQA models reveals significant limitations in existing methods for RGC evaluation. The wide MOS distributions and inconsistent model performance across categories highlight the need for RGC-tailored evaluation frameworks.These results demonstrate that no existing VQA model can be extensively reliable to assess RGC quality, necessitating new approaches that account for machine-specific motion patterns, egocentric perspectives, and device-induced artifacts. Our work establishes the first foundation for RGC quality research, and is of crucial importance to applications like surveillance, teleoperation, entertainment. We hope that the proposed RGC database will promote the development of in-depth research works on robotics and RGC video quality assessment.


\bibliographystyle{ACM-Reference-Format}
\bibliography{sample-base}




\end{document}